\documentclass[letterpaper]{article} 
\usepackage{aaai23-r2hcai}  
\usepackage{times}  
\usepackage{helvet}  
\usepackage{courier}  
\usepackage[hyphens]{url}  
\usepackage{graphicx} 
\def\UrlFont{\rm}  
\usepackage{natbib}  
\usepackage{caption} 
\usepackage{algorithm}
\usepackage{algorithmic}
\usepackage{subcaption}
\usepackage{graphicx}
\usepackage{mathtools}
\usepackage{amsmath,amssymb}
\usepackage{breqn}
\usepackage{physics}

\newtheorem{assumption}{Assumption}
\newtheorem{definition}{Definition}
\usepackage{newfloat}
\usepackage{listings}
\DeclareCaptionStyle{ruled}{labelfont=normalfont,labelsep=colon,strut=off} 
\floatstyle{ruled}
\newfloat{listing}{tb}{lst}{}
\floatname{listing}{Listing}
\usepackage{fancyhdr}
\fancypagestyle{firstpagehf}
{
   \fancyhf{}
   \fancyhead[HC]{The AAAI 2023 Workshop on Representation Learning for Responsible Human-Centric AI (R$^2$HCAI)}
   \fancyfoot[C]{\thepage}
}

\title{Exploiting Unlabeled Data for Feedback Efficient Human Preference based Reinforcement Learning}
\author{
   Mudit Verma\textsuperscript{\rm 1} \quad
   Siddhant Bhambri \textsuperscript{\rm 1} \quad
   Subbarao Kambhampati \textsuperscript{\rm 1} \quad
}
\affiliations{
    \textsuperscript{\rm 1} SCAI, Arizona State University\\
    muditverma@asu.edu, siddhantbhambri@asu.edu, rao@asu.edu
    
}

\begin{document}
\thispagestyle{firstpagehf}
\maketitle

\begin{abstract}
Preference Based Reinforcement Learning has shown much promise for utilizing human binary feedback on queried trajectory pairs to recover the underlying reward model of the Human in the Loop (HiL). While works have attempted to better utilize the queries made to the human, in this work we make two observations about the unlabeled trajectories collected by the agent and propose two corresponding loss functions that ensure participation of unlabeled trajectories in the reward learning process, and structure the embedding space of the reward model such that it reflects the structure of state space with respect to action distances. We validate the proposed method on one locomotion domain and one robotic manipulation task and compare with the state-of-the-art baseline PEBBLE. We further present an ablation of the proposed loss components across both the domains and find that not only each of the loss components perform better than the baseline, but the synergic combination of the two has much better reward recovery and human feedback sample efficiency.
\end{abstract}

\section{Introduction}
\label{sec:introduction}
Reinforcement Learning (RL), especially Deep Reinforcement Learning has gained immense popularity with significant leaps in allowing agents to learn complex behaviors, in high dimensional state and action spaces \cite{mnih2015human, arulkumaran2017deep}. However, much of the successes have also been attributed to well specified reward functions which ground the agent's behavior and subsequent task in the expected manner. As prior works have argued, the specification of low level reward functions for seemingly easy tasks could be quite difficult and may still result in inexplicable and unexpected results \cite{verma2019making, verma2021perfect, gopalakrishnan2021computing, gopalakrishnan2021synthesizing} potentially affecting trust between Human-AI \cite{zahedi2021trust, zahedi2022modeling}. For example, works like \cite{reward-hacking-init, reward-hacking-example} have raised the issues of reward hacking and reward exploitation where the RL agents would discover behaviors that seems to be ``cheating" or incorrect and yet maximize the expected cumulative reward. This has also gotten attention from the explainable AI community where they attempt to analyze whether the agent is actually behaving in the intended manner \cite{adv-conf, blackbox, lingua-franca}.  In recent literature, a potential solution to such issues has been to allow a human in the loop (HiL) to specify their preferences as feedback on queried trajectory pairs \cite{wilson2012, christiano, prior} instead of feeding hand-designed reward functions to the system. 

Recent Preference-based Reinforcement Learning (PbRL) methods like \cite{pebble, surf, prior, christiano, soni2022towards} can efficiently utilize the queries made to the human in the loop via several key ideas like pre-training \cite{pebble}, improved query sampling strategy \cite{pebble}, data augmentation \cite{surf, expand, guan2020explanation}, or priors specified over reward function \cite{prior}. In this work, we present a complementary approach by which we can exploit unlabeled trajectories for improved reward recovery of the underlying human reward function. Our method relies on two main observations: the first is that an extremely large population of trajectories lie in the agent's buffer (collected over training episodes) that are not used in the reward learning process and that the representation space for the reward function being learnt is not reflective of how the state space is structured. Specifically, our first observation reinforces the fact that much of the explored trajectories do not actually participate in the reward learning process and in fact their best change to affect the reward function is once they get sampled and queried to the human in the loop. We posit that this untapped data source can greatly improve reward recovery and reduce the feedback sample complexity. Our second observation notes that the reward function being learnt may not conform to the structure of the state space simply because it doesn't get exposed to as many data points (in comparison to, say, the policy approximation function). We utilize our observations to improve performance of RL agents in recovering the underlying reward function and learn a good policy by exploiting the rich unlabeled trajectory data. Although works like SURF \cite{surf} have proposed a semi-supervised learning approach to utilize unlabeled trajectory data, they would generate labels for unlabeled trajectories and use these data points as if they were given by the human in the loop. We argue that this is still an indirect way of updating the reward function via these unlabeled trajectories since all the information about the predicted reward values is lost when providing a hard decision label on which trajectory was preferred. Having said this, the proposed triplet loss, in part, attempts to perform semi-supervised \cite{yantian_triplet_loss, surf} learning and builds over SURF for PbRL. Finally, empirically we found that our method easily outperforms SURF on both the locomotion and robotic manipulation task.

\section{Background}
\label{sec:background}
Reinforcement learning allows for agents interacting in an environment $\mathcal{E}$ where at each discrete timestep $t$, the agent receives an observation $o_t$ from the environment and chooses an action $a_t$ based on its policy $\pi$. As in conventional RL frameworks we assume that the underlying system is a Markov Decision Process, i.e. the tuple $<\mathcal{S, T, A, \tilde{R}}_h, \gamma>$ describing the state space $\mathcal{S}$, agent's action space $\mathcal{A}$, the underlying environment transition dynamics $\mathcal{T}$, the discount factor $\gamma$ where the agent's goal is to maximize the return $\sum_{k=0}^{\infty}\gamma^k \mathcal{\tilde{R}}_h(s_{t+k}, a_{t+k})$ computed over the reward system $\mathcal{R}_h$ in concern. In the preference based reinforcement learning setup we are interested in, the goal of the agent is two fold, first to infer the human's underlying reward model $\mathcal{\tilde{R}}_h$ via binary feedback over trajectory pairs and further use the learnt reward model $R_h$ to compute a policy $\pi_\phi$ parameterized by $\phi$ to maximize discounted cumulative return over $R_h$. 

We utilize the formulation presented in \cite{wilson2012} for the preference based reinforcement learning problem where the agent queries the human in the loop with a trajectory pair $\tau_0, \tau_1$, $\tau_i = \{(s_k, a_k), (s_{k+1}, a_{k+1} \cdots (s_{k+H}, a_{k+H}))\}$ for a binary feedback $y \in \{0,1\}$ indicating their preferred trajectory. Such feedbacks along with the queried trajectories are stored in a dataset $D_\tau$ as tuples $(\tau_0, \tau_1, y)$. Following the Bradley Terry model \cite{bradley-terry} to compute probability of one trajectory be preferred over another, recent line of works like \cite{pebble, christiano} approximates the human reward function as $R_h$, parameterized by, say, $\psi$, by solving a supervised learning problem where the returns computed over the learnt reward function are higher for trajectories that were preferred by the human in the loop than the returns computed on the non-preferred trajectory. This is done by minimizing the cross-entropy between the predictions and ground truth human labels as follows: 
\begin{dmath}
    \mathcal{L}_{CE}=-\displaystyle \mathop{\mathbb{E}}_{(\tau_0, \tau_1, y)\sim \mathcal{D}}[y(0)\text{log}P_\psi[\tau_0 \succ \tau_1] + {y(1)\text{log}P_\psi[\tau_1 \succ \tau_0]]}
\end{dmath}

where probabilities $P_\psi$ are computed using the approximated reward function $R_h$ as : 
 \begin{equation}
     P_{\psi}[\tau_0 \succ \tau_1] = \frac{\exp(\sum_{t}R_h(s_t^0, a_t^0))}{\sum_{i \in \{0,1\}} \exp(\sum_{t}R_h(s_t^i, a_t^i))}
 \end{equation}

For our experiments we use the PEBBLE \cite{pebble} as the backbone, however the proposed work's implications are not limited to PEBBLE and can be applied to any PbRL method that attempts to approximate the underlying human reward function. In fact, our method is complementary to existing methods that typically improve the agent's performance or reward recovery by improved query sampling strategy or data augmentation to name a few.

\section{Method}
We operationalize our observations as mentioned in the Introduction section by utilizing the unlabeled trajectory data and propose a solution to leverage each of the two observations. The Result section shows that these two insights are actually complementary. For our first observation, there exists a rich source of unlabeled data and that it could be helpful to ensure that the reward function being learnt is affected by it, we make the following assumption : 
\begin{assumption} \label{asmp:1}
A trajectory $\tau$, sampled under a policy $\pi_{\phi}$, that has not been queried to the human in the loop is assumed to be preferred by the human.
\end{assumption}

Since there exist a large bank of trajectories that has not been queried to the HiL, Assumption \ref{asmp:1} makes a paternalistic choice about whether those trajectories would be preferred by the HiL over some other trajectory. Moreover, we can use this assumption to ensure that the reward model can now use these unlabeled trajectories. In contrast to prior works like \cite{surf} that extract labels via learnt reward model over the unlabeled trajectory data, we use our assumption \ref{asmp:1} and propose a triplet loss that directly updates the reward model as follows,

\begin{dmath}
    \mathcal{L}^t(\tau; D_h) = {\frac{1}{|D_h|}\sum_{\tau_g, \tau_b \sim D_h} max(0, ||\vectorbold{R}(\tau) - \vectorbold{R}(\tau_g)||^2} - {||\vectorbold{R}(\tau) - \vectorbold{R}(\tau_b)|| + m)}
\end{dmath}

where $m$ is the margin hyperparameter. We overload the notation for reward to reflect the rewards for the trajectory states as a vector, i.e. $\vectorbold{R(\tau)} = \begin{bmatrix} R(s_0) & R(s_1) & \cdots & R(s_{T-1}) & R(s_{T})) \end{bmatrix}^T$ for a trajectory $\tau$ of length $T$. $D_\tau$ is the bank of trajectories sampled by the agent, a common element used by off-policy RL algorithms as the replay buffer. $D_h$ is the dataset of preference labels over the queried trajectory pairs where we use $\tau_g$ to denote the trajectory that was preferred over $\tau_b$. The triplet loss uses the preferred trajectories as the positive samples, the dis-preferred trajectories as the negative sample and the unlabeled trajectory $\tau$ as the anchor.

For our second observation, we propose an action based loss (variants of which have been seen in works like \cite{markovchain_distance, action_distance_sorb}) for the reward model to impose a soft constraint on the state embedding being learnt by the reward model in an attempt to ensure that the reward model also reflects the structure of state space (with respect to action distances). 

\begin{definition}
Action distance $A_d$ between two states under some policy $\pi_{\phi}(s)$ and transition dynamics $\mathcal{T}(s,a,s')$ is given by the expected number of action steps taken to reach a state $s_2$ from $s_1$.
\end{definition}

We propose to enforce such a soft constraint in the embedding space of the reward model, $R_{e}(s)$ computes the embedding of the state $s$, by ensuring that the euclidean distance between the embedding of two states $s_1$ and $s_2$ reflects the action distance $A_d(s_1, s_2)$. This can be achieved by minimizing the Mean Squared Error (MSE) between the computed distance in the embedding space and the action distance as follows : 
\begin{equation}
    \mathcal{L}^a(D_p) = \frac{1}{|D_p|}\sum_{s_i, s_j, d_y \sim D_p} (||R_e(s_i) - R_e(s_j)||^2 - d_y)^2
\end{equation}

where for a $s_i, s_j$ are pair of states in the dataset $D_p = (s_i, s_j, d_y)$ which consists of the computed ground truth action distances between them as $d_y$. What remains is to create this dataset $D_p$. We utilize the trajectory bank $D_\tau$ (and also include trajectories in $D_h$) to obtain $D_p$. The key idea is that since the action distance ground truth that we want is an expectation over number of actions taken to reach $s_j$ from $s_i$, we can approximate this action distance by sampling a state $s_i, s_j \in \tau$  where $j > i, \tau \in D_\tau$ and use the number of action steps taken in the trajectory from $s_1$ to $s_2$ as the ground truth distance $d_y = j - i$.  An important note is that the distances $d_y$ in the dataset $D_p$ should be from the agent's current policy $\pi_{\phi}$. For off-policy RL algorithms where the replay buffer, $D_\tau$, would contain trajectories sampled from a stale policy, we emulate the required behavior of the dataset $D_p$ by ensuring that only the last $k$ trajectories added to the dataset $D_\tau$ are used to compute $D_p$.

We finally utilize a linear combination of the two proposed loss functions $\mathcal{L}^t$, triplet loss, and $\mathcal{L}^a$, action distance loss, with the cross entropy loss $L_{CE}$ (see Section \ref{sec:background}) to update the reward model as : 
\begin{dmath}
    \mathcal{L}^{reward} = {\lambda_{CE}\mathcal{L}^{CE}(D_h) + \lambda_{t}\mathcal{L}^t(D_\tau)} + {\lambda_{a}\mathcal{L}^a(D_p(D_\tau[k..T]))}
\end{dmath}

where $\mathcal{L}^{CE}$ is computed over $D_h$ that contains the queried trajectory pairs with human binary feedbacks (mean over the samples), $\mathcal{L}^t$ is computed over the unlabeled trajectory buffer (mean over all the trajectories) $D_\tau$ and $\mathcal{L}^a$ is computed over the dataset of state pairs with action distance $D_p$ (mean over all the tuples) created from the $k$ most recent trajectories added to $D_\tau$.

\section{Experiments}
\label{sec:result}

\begin{figure}
     \centering
     \begin{subfigure}[b]{0.47\textwidth}
         \centering
         \includegraphics[width=\textwidth]{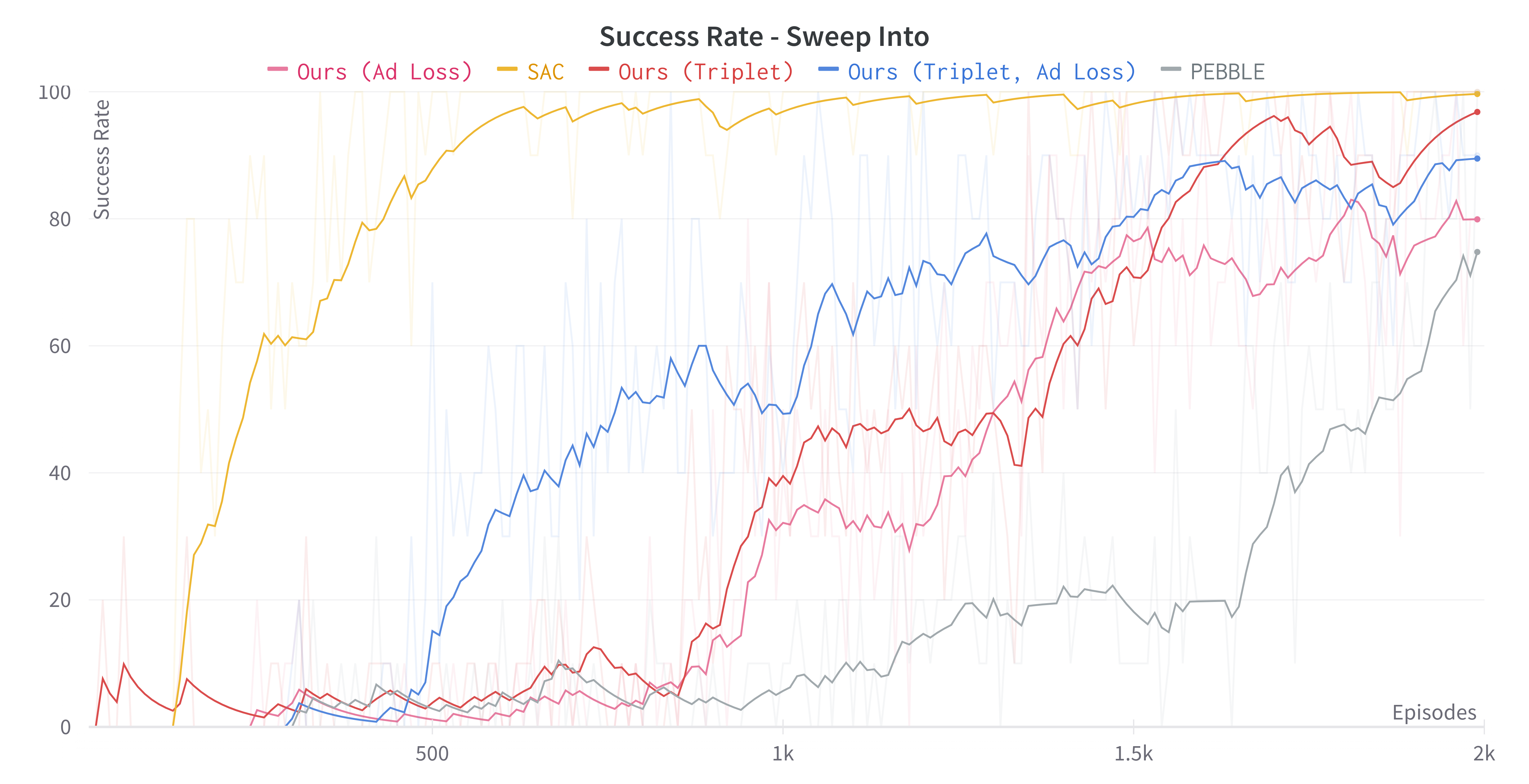}
         \caption{Success Rate}
         \label{fig:success_sweep}
     \end{subfigure}
     \begin{subfigure}[b]{0.47\textwidth}
         \centering
         \includegraphics[width=\textwidth]{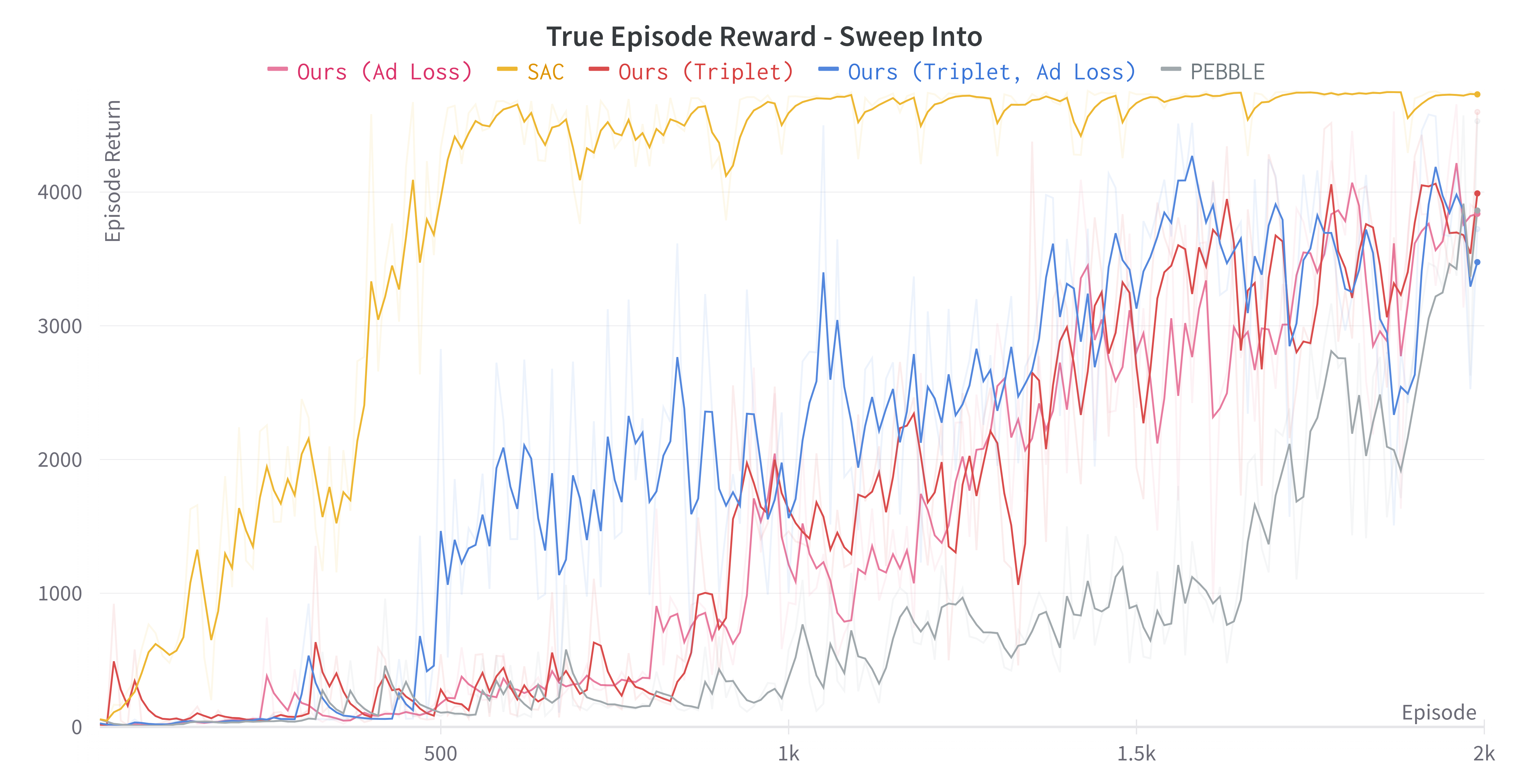}
         \caption{Return of learnt $\pi_\phi$ on ground truth reward $R_h$}
         \label{fig:reward_sweep}
     \end{subfigure}
        \caption{Evaluation curves on the robotic manipulation task of Sweep-Into as measured on the success rate and the ground truth human reward $R_h$.}
        \label{fig:sweep}
\end{figure}

We wanted to investigate the following two questions via empirical evaluations : 
\begin{enumerate}
    \item Do the proposed losses improve the existing state of the art in preference based RL in terms of reward recovery, feedback efficiency and performance of the learnt policy?
    \item Are the two losses proposed in this work complementary, and more so synergic?
\end{enumerate}

We validated our proposed method via experiments on two domains, of which one is a locomotion task (Quadruped) and the other is a robot manipulation task (Sweep-Into). Recent literature \cite{pebble, surf} on Preference-based Reinforcement Learning has showcased results particularly for continuous control tasks like locomotion and robot manipulation. Future work involves more extensive evaluation on other tasks like Walker, Cheetah, Drawer Open, Window Open etc.,  and further investigation of the method's benefits on explicit knowledge tasks in discrete action space domains like Montezuma's Revenge and MS-Pacman.

In order to systematically compare our work with the baselines we use a synthetic oracle that has a fixed reward model for the agent ($R_h$). A good PbRL algorithm should be able to recover this reward model, ${R_\psi}$, and subsequently a policy learnt on the recovered reward model, $\pi_\phi$, when evaluated on the human's reward model, should yield high expected return. We follow existing literature and assume the environment's original reward function as the oracle's underlying ground truth reward model $R_h$ and provide a feedback label as follows : 
\begin{equation}
\label{eq:oracle_label}
    y(\tau_0, \tau_1) = \begin{cases}
    0 & \sum_{i} R_h(\tau_0) > \sum_{i} R_h(\tau_1) \\ 
    1 & \sum_{i} R_h(\tau_0) < \sum_{i} R_h(\tau_1) \\ 
    \end{cases}
\end{equation}

Note that there can exist a third case in equation \ref{eq:oracle_label}, where the human has exactly equal preference over the trajectory pairs, and does not occur frequently after a few updates to the reward model. Even though prior works have highlighted the use of soft cross entropy loss (instead of $\mathcal{L}^{CE}$) to handle such situations it is not central to the problem of PbRL with sparse equal preferences over a large number of randomly sampled trajectory pairs, and we do not expect the use of $\mathcal{L}^{CE}$ to limit the presented losses $\mathcal{L}^t, \mathcal{L}^a$ and concepts in any way. We do, however, plan to investigate the implications of the proposed techniques when the human in the loop is allowed to mark trajectory queries as equally preferred.

\subsection{Implementation Details : } The underlying RL algorithm used to train the policy is SAC \cite{haarnoja2018soft} for baseline PEBBLE and ours. For all our experiments we query feedback over trajectories of a fixed length of 50 and borrow other hyperparameters used for the baseline algorithm, SAC, and (for the hyperparameters that are common) ours from \cite{bpref}. We use $\lambda_CE = 1, \lambda_t = 0.5, \lambda_a = 3$ to ensure that all the computed loss values have a similar scale. To compare with the original baseline results, we set the max feedback threshold to 1000 for Quadruped-Walk and 10000 for Sweep-Into. We use the same architecture for the policy and the reward models as in \cite{pebble}, and use the penultimate layer in the reward model as the embedding space required by the action distance loss.

\subsection{Results}
For both of the environments we compare our method against two baselines, PEBBLE \cite{pebble} that follows a similar training paradigm but only uses the Cross Entropy Loss and an RL (SAC) baseline that has access to the underlying ground truth reward $R_h$. Our results for the baseline PEBBLE and SAC performance on these domains can be corroborated as shown by \cite{surf}. 

\begin{figure}
    \centering
    \includegraphics[width=0.45\textwidth]{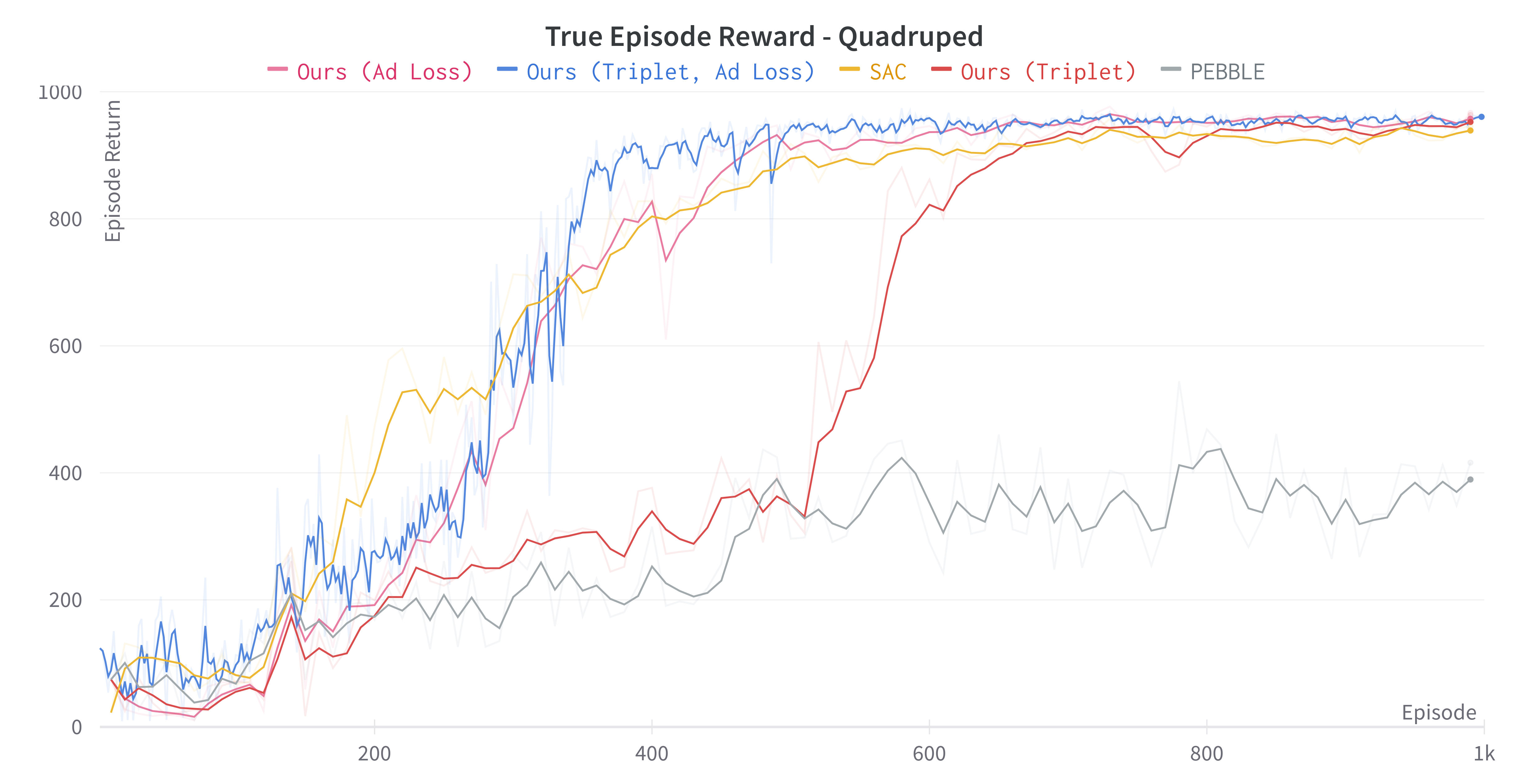}
    \caption{Evaluation curves on the locomotion task of Quadruped as measured on the ground truth human reward $R_h$.}
    \label{fig:quad}
\end{figure}

For the task of Quadruped-Walk, PEBBLE (see Fig. \ref{fig:quad}) only achieves performance levels of a return of $\sim$400 which is much less than that of our SAC baseline $\sim$980, in contrast, Our work (Triplet Loss, Ad Loss) performs significantly better than the PEBBLE baseline and reaches performance levels of SAC (even outperforming it for a while, that has potentially been attributed to better exploration in PbRL setups than SAC) with significantly fewer feedback samples. Similar gains are seen for the task of Sweep-Into (see Fig. \ref{fig:sweep}) where PEBBLE utilizes a very high number of feedback samples to show any improvements in the success rate (Fig. \ref{fig:success_sweep}) and the return over ground truth reward (Fig. \ref{fig:reward_sweep}). The combination of triplet loss and action distance loss easily outperforms PEBBLE baseline and reaches a reasonable success rate and return values.

We also compare our results with SURF, and refer the readers to the reported results in \cite{surf} for the locomotion and the robotic manipulation task. In the reported results, for Sweep-Into, SURF performs closely to our baseline PEBBLE and achieves $\sim 75\%$ success rate with the same number of feedbacks compared to almost perfect $100\%$ for our work. Similarly, SURF's doesn't show any gains on Quadruped over PEBBLE in their reported results and, as discussed, the learnt policy achieves $50\%$ of the expected return by SAC, whereas as shown in fig. \ref{fig:quad}, we not only show significant improvement over the baseline but also achieve similar performance as SAC.

Finally, to ascertain that the two losses presented in this work that utilize unlabeled trajectory data (triplet loss $\mathcal{L}^t$ and action distance loss $\mathcal{L}^a$) work complementary to each other, we perform an ablation comparing the two losses together against only the triplet loss and only the action distance loss. We find that, although, in both the environments even only one of the losses easily outperform the baseline PEBBLE, the synergic combination of the two increases the performance measures substantially with much fewer human feedback samples. 


\section{Discussion}

In this work we presented two key observations (and corresponding loss functions) regarding the utilization of unlabeled trajectories for a PbRL agent. We first proposed a triplet loss under the optimistic assumption that an unlabeled trajectory would be preferred by the human, and secondly, our action distance loss function that attempts to structure the embedding space of the reward model being learnt to reflect action distances between state pairs. We show that although these individual losses perform much better than the baseline PbRL and RL (SAC) in terms of reward recovery and human feedback sample efficiency, the synergic combination of these yield a more powerful PbRL agent with low demands of human sample feedback and high performance. 

Future work includes a more thorough investigation of the effects of proposed method across diverse locomotion, robotic manipulation as well as explicit knowledge discrete domains. We also intend to perform an extensive evaluation of the effects of several key hyperparameters like the query trajectory length and maximum number of feedbacks to further bolster our claims.

\section*{Acknowledgements}
Kambhampati's research is supported by the J.P. Morgan Faculty Research Award, ONR grants N00014-16-1-2892, N00014-18-1-2442, N00014-18-1-2840, N00014-9-1-2119, AFOSR grant FA9550-18-1-0067 and DARPA SAIL-ON grant W911NF19-2-0006. 

\bibliography{aaai23} 
\end{document}